\documentclass[10pt,twocolumn,letterpaper]{article}

\usepackage[pagenumbers]{cvpr}      

\usepackage{multirow}
\usepackage{booktabs} 
\usepackage{graphicx} 
\usepackage{xcolor}   

\usepackage{array}    
 \usepackage{tikz}
\usetikzlibrary{arrows.meta,calc}
\usepackage{graphicx,amsmath,amssymb}

\usepackage{stfloats}
\usepackage[table]{xcolor}

\definecolor{cvprblue}{rgb}{0.21,0.49,0.74}
\usepackage[pagebackref,breaklinks,colorlinks,allcolors=cvprblue]{hyperref}

\def\confName{CVPR}
\def\confYear{2027}

\title{Self-Supervised Dataset Distillation via Mutual Information Maximization}

\author{
Mohammad Hossein Moslemi$^{1,4}$ \quad
Mahdi Beitollahi$^{2}$ \quad
Zhimin Mei$^{1}$ \quad
Jiale Cai$^{1}$\\
Bissan Ghaddar$^{1,3}$ \quad
Boyu Wang$^{1,4}$\\[0.4em]
$^{1}$Western University, London \qquad
$^{2}$Queen's University, Kingston\\
$^{3}$Ivey Business School, London \qquad
$^{4}$Vector Institute, Toronto\\[0.4em]
{\tt\small mohammad.moslemi@uwo.ca \quad Beitollahi.mahdi@queensu.ca}\\
{\tt\small \{zmei22,jcai336\}@uwo.ca \quad bghaddar@ivey.uwo.ca \quad bwang@csd.uwo.ca}
}

\begin{document}
\maketitle

\begin{abstract}


Dataset distillation (DD) aims to compress a large dataset into a small synthetic one that remains effective for training neural networks.
Despite its success in supervised learning, DD remains underexplored in self-supervised learning (SSL), where distilled data must preserve the information needed to learn representations that remain consistent across augmentations.
To bridge this gap, we introduce \textbf{M}utual \textbf{I}nformation \textbf{P}rototype \textbf{A}lignment (MIPA), an efficient SSL dataset distillation approach that optimizes synthetic images to preserve augmentation-invariant information.
Specifically, MIPA treats synthetic images as learnable prototypes and uses a frozen encoder to assign real samples to them based on feature similarity.
A synthetic dataset that preserves such augmentation invariance should yield stable prototype assignments across multiple augmented views of the same real image.
MIPA encourages this by maximizing mutual information between cross-view assignments while avoiding trivial solutions in which all real samples are assigned to only a few synthetic prototypes.
Evaluations on image classification benchmarks show that synthetic data generated by MIPA yield higher-quality representations than existing distillation approaches while reducing time and storage demands during distillation.
Code is available at \href{https://github.com/mhmoslemi/MIPA.git}{https://github.com/mhmoslemi/MIPA.git}

\end{abstract}

\section{Introduction}
\label{sec:intro}

Dataset distillation (DD) seeks to condense a large dataset into a compact set of synthetic samples that approximate the effect of training on the full dataset while reducing memory and computational requirements \cite{wang2018dataset}. This makes DD attractive for edge model training and continual learning, while its privacy implications have also received attention \cite{zhao2025does, sajedi2023datadam, cazenavette2022dataset,dong2022privacy}. However, most existing DD methods focus on supervised settings that assume access to class labels, which are often scarce or unavailable in practice. In such cases, self-supervised learning (SSL) offers an alternative by pretraining an encoder on large amounts of unlabeled data, which can subsequently be adapted to a range of downstream tasks.
Accordingly, the computational and memory advantages of DD could also benefit SSL. Despite this potential, SSL-DD remains relatively underexplored.

A compact synthetic dataset for SSL must preserve sufficient augmentation-invariant information from the original large-scale one to support learning useful representations with SSL.
Existing supervised DD methods, including gradient matching~\cite{zhao2020dataset, zhao2021dataset}, distribution matching~\cite{zhao2023dataset}, and trajectory matching~\cite{cazenavette2022dataset}, are not effective in this setting as their objectives rely on class-label supervision rather than the specific requirements of SSL.
More recently, KRR-ST~\cite{lee2024self} and MKDT~\cite{joshi2025dataset} have attempted to bridge this gap by adapting supervised DD techniques 
using auxiliary target representations as supervision signals
without fully accounting for the distinct learning objectives of SSL.
Consequently,
both methods require auxiliary target representations
alongside the synthetic images during SSL pretraining.
This dependence undermines the goal of standalone DD by increasing storage overhead, 
requiring target-matching losses
instead of standard SSL objectives, and introducing potential security vulnerabilities associated with malicious teacher representations~\cite{CHEN2024104041}.
Notably, we observe that removing these auxiliary representations and pretraining only on synthetic images does not reliably outperform randomly selecting real samples from the original dataset.

\begin{figure}[h]
    \centering
    \begin{subfigure}[b]{.33\columnwidth}
        \centering
        \includegraphics[width=\linewidth]{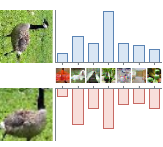}
        \caption{}
        \label{fig:assignment-distributions}
    \end{subfigure}\hfill
    \begin{subfigure}[b]{.65\columnwidth}
        \centering
        \includegraphics[width=\linewidth]{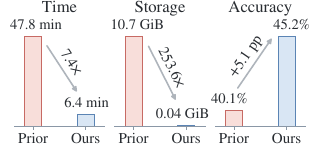}
        \caption{}
        \label{fig:efficiency-comparison}
    \end{subfigure}
\caption{
\textbf{Assignment distributions and efficiency comparison.}
Figure~\ref{fig:assignment-distributions} shows two augmented views of the same real image (left) mapped to shared synthetic prototypes (middle); the top and mirrored bottom bars indicate the assignment distributions.
Figure~\ref{fig:efficiency-comparison} compares Prior (MKDT~\cite{joshi2025dataset}) and Ours (MIPA) on CIFAR-10 with a synthetic set of 0.2\% of the training data. We pretrain ResNet-18 from scratch with SimCLR on synthetic images only, followed by a linear classifier with 1\% labeled data. 
}
    \label{fig:assignment-efficiency}
\end{figure}

In this work, rather than adapting supervised DD to SSL, we design an approach that directly accounts for the augmentation-invariance requirement of SSL.
To this end, we introduce \textbf{M}utual \textbf{I}nformation \textbf{P}rototype \textbf{A}lignment (MIPA), which treats synthetic images as learnable prototypes in the representation space.
In particular, for each real image, MIPA computes a probability distribution over these prototypes based on feature similarity, indicating the degree to which each synthetic image matches the real sample. It then evaluates these assignments over multiple views of the same image.
Preserving augmentation-invariant information in synthetic data should yield consistent assignments for different views of the same real image.
Simply minimizing distance-based objectives between assignments can result in trivial solutions in which all real images are assigned to the same limited subset of synthetic prototypes.
Instead, MIPA maximizes mutual information between cross-view assignments by optimizing the synthetic data, promoting consistency in assignments while avoiding such collapse, as demonstrated by our analysis and empirical results.
Figure~\ref{fig:assignment-distributions} illustrates this procedure by showing assignment distributions from two views of the same real image.

We evaluate MIPA on image classification benchmarks across multiple architectures and SSL methods. Our results indicate that MIPA yields higher-quality representations than prior distillation and sample selection approaches. More importantly, as shown in Figure~\ref{fig:efficiency-comparison}, MIPA achieves higher accuracy while significantly reducing the time and storage required during distillation.
In particular, we make the following contributions:
\begin{enumerate}[label=\arabic*., start=1]




\item We identify augmentation consistency and prototype diversity as key to SSL distillation, demonstrating that consistency alone can lead to trivial solutions.





\item We propose MIPA, which maximizes mutual information between cross-view prototype assignments to enable SSL pretraining from scratch using only synthetic data.

    \item Comprehensive experiments confirm MIPA's effectiveness on downstream tasks while substantially reducing distillation time and storage requirements.
\end{enumerate}

\section{Related Work}

\subsection{Dataset Distillation}

Dataset distillation was first formulated as a bilevel optimization problem~\cite{wang2018dataset}. Subsequent approaches include gradient matching~\cite{zhao2020dataset}, distribution matching~\cite{zhao2023dataset}, kernel-based meta-learning~\cite{nguyen2020dataset}, and training trajectory matching~\cite{cazenavette2022dataset,du2023minimizing}. Recent work further incorporates optimal transport into distribution-matching objectives~\cite{liu2025dataset}. Complementary studies investigate adversarial attacks~\cite{yang2025sneakdoor}, privacy~\cite{zhao2025does}, fairness~\cite{moslemi2026fair}, and robustness to domain shifts~\cite{loo2024large,vahidian2025group}. Most of these methods operate in supervised settings. Other approaches leverage pretrained SSL encoders for data distillation under different assumptions. SC-DD~\cite{zhou2024scdd} reconstructs images using batch-normalization statistics with class supervision, while LGM~\cite{cazenavette2025pretrained} learns labeled synthetic data by matching linear-classifier gradients on frozen SSL features. SRG~\cite{li2026self} further utilizes class prototypes in SSL representation space to guide diffusion-based synthesis.

Self-supervised dataset distillation instead seeks to compress data without any class labels.
KRR-ST~\cite{lee2024self} adapts kernel-based matching to this setting. However, \textsc{MKDT}~\cite{joshi2025dataset} reported that KRR-ST did not consistently outperform SSL pretraining on random real subsets and further showed that MTT~\cite{cazenavette2022dataset}, although label-agnostic in principle, performs poorly when applied to SSL. \textsc{MKDT} instead replaces high-variance SSL trajectories with lower-variance knowledge-distillation trajectories. \citet{yu2025boost} further investigate SSL distillation by jointly optimizing synthetic images and target representations, using predefined augmentations and an approximation network to mitigate augmentation-related instability. Similar to KRR-ST and MKDT, this approach retains auxiliary representation targets for subsequent training. In contrast, \textsc{MIPA} uses a frozen SSL encoder solely to guide distillation, avoids inner-loop encoder training, and produces synthetic images that independently support SSL pretraining from scratch without class labels or auxiliary target representations.

\subsection{Self-Supervised Learning}
Self-supervised learning aims to learn transferable representations from unlabeled data. 
Contrastive methods such as SimCLR~\cite{chen2020simple} and MoCo~\cite{he2020momentum} align augmented views of the same image while separating representations of different images.
BYOL~\cite{grill2020bootstrap} and SimSiam~\cite{chen2021exploring} learn through prediction between views without explicit negative pairs. Barlow Twins~\cite{zbontar2021barlow} regularizes representation statistics by encouraging the cross-correlation matrix computed between two views toward the identity matrix.
Clustering-based methods such as SwAV~\cite{caron2020unsupervised} instead assign representations of augmented views to learnable prototypes and enforce consistency between their assignments across views.
Masked prediction techniques learn from partially observed inputs. For example, MAE~\cite{he2022masked} masks random patches of the input image and trains an encoder-decoder architecture to reconstruct the missing pixels. In contrast, I-JEPA~\cite{assran2023self} predicts the representations of target regions from a context region of the same image directly in feature space. 
In this work, we focus primarily on SimCLR and conduct additional evaluations using Barlow Twins.

\section{Preliminaries}
\subsection{Problem Statement}
\label{sec:problem-formulation}

Self-supervised dataset distillation aims to construct a small synthetic dataset $\mathcal{S} = \{x_j^s\}_{j=1}^{m}$ from a large unlabeled dataset $\mathcal{T} = \{x_i^t\}_{i=1}^{n}$, where $m \ll n$, such that an encoder pretrained on $\mathcal{S}$ using self-supervised learning (SSL) achieves downstream performance comparable to an encoder pretrained on $\mathcal{T}$ \cite{joshi2025dataset, lee2024self}. Class labels are unavailable during both distillation and pretraining.

\subsection{Self-Supervised Learning}
\label{sec:ssl-preliminaries}

Self-supervised learning uses training signals derived from unlabeled inputs to learn representations. Let $f_{\theta}$ denote an encoder mapping images from the input space $\mathcal{X}$ to $p$-dimensional representations in $\mathbb{R}^{p}$. A projection head $g_{\phi}$ maps these representations to $q$-dimensional vectors used to compute the SSL loss. The encoder and projection-head parameters are $\theta$ and $\phi$, with joint parameters $\omega = (\theta,\phi)$. Given a batch $\mathcal{B} = \{x_i\}_{i=1}^{B}$ of $B$ images, two augmentations $a_i^{(1)},a_i^{(2)} \sim \mathcal{A}$ are independently sampled for each image, where $\mathcal{A}$ denotes the augmentation distribution. The resulting views have projected representations $z_i^{(v)} = g_{\phi}(f_{\theta}(a_i^{(v)}(x_i)))$, where $v \in \{1,2\}$ indexes the two views.
SimCLR~\cite{chen2020simple} encourages higher cosine similarity between $z_i^{(1)}$ and $z_i^{(2)}$ than between representations of different images in the batch. Barlow Twins~\cite{zbontar2021barlow} instead computes the cross-correlation matrix between the two sets of projected representations after standardizing each dimension across the batch.

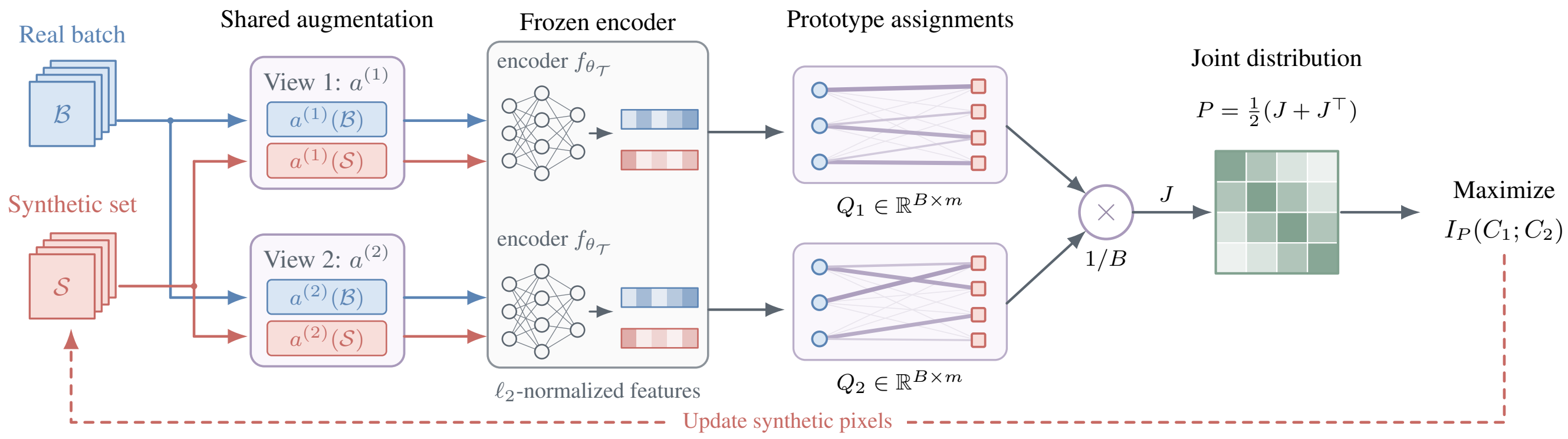
\begin{figure*}[t!]
\centering
\resizebox{\linewidth}{!}{%
\begin{tikzpicture}[x=0.80cm,y=0.80cm,
    >=Latex,
    line cap=round,line join=round,
    every node/.style={inner sep=0pt},
    stage/.style={font=\normalfont\small},
    textlabel/.style={font=\normalfont\small},
    note/.style={font=\normalfont\footnotesize,text=ddgray}]

\definecolor{ddblue}{RGB}{91,132,180}
\definecolor{ddbluefill}{RGB}{217,230,244}
\definecolor{ddred}{RGB}{199,106,100}
\definecolor{ddredfill}{RGB}{248,222,218}
\definecolor{ddpurple}{RGB}{169,154,187}
\definecolor{ddpurplefill}{RGB}{250,247,252}
\definecolor{ddgreen}{RGB}{102,142,120}
\definecolor{ddgreenfill}{RGB}{247,250,243}
\definecolor{ddgray}{RGB}{91,101,111}

\node[stage] at (5.77,6.82) {Shared augmentation};
\node[stage] at (10.02,6.82) {Frozen encoder};

\node[textlabel,text=ddblue] at (1.77,6.62) {Real batch};
\foreach \xx/\yy in {1.43/5.19,1.32/5.08,1.21/4.97,1.10/4.86}{%
  \draw[fill=ddbluefill,draw=ddblue,line width=.70pt]
    (\xx,\yy) rectangle ++(1.02,1.01);
}
\node[font=\small,text=ddblue] at (1.61,5.365) {$\mathcal{B}$};
\node[textlabel,text=ddred] at (1.77,3.91) {Synthetic set};

\foreach \xx/\yy in {1.43/2.48,1.32/2.37,1.21/2.26,1.10/2.15}{%
  \draw[fill=ddredfill,draw=ddred,line width=.70pt]
    (\xx,\yy) rectangle ++(1.02,1.01);
}
\node[font=\small,text=ddred] at (1.61,2.655) {$\mathcal{S}$};

\draw[->,draw=ddblue,line width=.95pt] (2.48,5.26)--(4.53,5.26);
\draw[draw=ddblue,line width=.95pt] (3.31,5.26)--(3.31,2.48);
\draw[->,draw=ddblue,line width=.95pt] (3.31,2.48)--(4.53,2.48);
\fill[ddblue] (3.31,5.26) circle (.050);

\draw[draw=ddred,line width=.95pt] (2.51,2.66)--(3.68,2.66);
\draw[draw=ddred,line width=.95pt] (3.68,1.87)--(3.68,4.62);
\draw[->,draw=ddred,line width=.95pt] (3.68,4.62)--(4.53,4.62);
\draw[->,draw=ddred,line width=.95pt] (3.68,1.87)--(4.53,1.87);
\fill[ddred] (3.68,2.66) circle (.050);

\begin{scope}[xshift=2.4mm]
\foreach \by/\v in {4.19/1,1.40/2}{%
  \draw[rounded corners=.17cm,draw=ddpurple,line width=.8pt,fill=ddpurplefill]
    (4.27,\by) rectangle (6.68,\by+2.07);
  \node[textlabel,text=ddgray] at (5.47,\by+1.73)
    {View \v: $a^{(\v)}$};
  \draw[rounded corners=.055cm,draw=ddblue,line width=.55pt,fill=ddbluefill]
    (4.53,\by+.82) rectangle (6.40,\by+1.36);
  \node[font=\footnotesize,text=ddblue] at (5.47,\by+1.09)
    {$a^{(\v)}(\mathcal{B})$};
  \draw[rounded corners=.055cm,draw=ddred,line width=.55pt,fill=ddredfill]
    (4.53,\by+.17) rectangle (6.40,\by+.71);
  \node[font=\footnotesize,text=ddred] at (5.47,\by+.44)
    {$a^{(\v)}(\mathcal{S})$};
}
\end{scope}

\foreach \yy in {5.26,2.48}{\draw[->,ddblue,line width=.95pt] (6.98,\yy)--(8.29,\yy);}
\foreach \yy in {4.62,1.87}{\draw[->,ddred,line width=.95pt] (6.98,\yy)--(8.29,\yy);}

\begin{scope}[xshift=5.2mm]
\draw[rounded corners=.16cm,draw=ddgray!70,fill=black!2,line width=.75pt]
  (7.64,1.38) rectangle (11.10,6.50);

\foreach \cy in {5.06,2.26}{%
  \node[note,anchor=south] at (8.7,\cy+.90) {encoder $f_{\theta_{\mathcal{T}}}$};
  \foreach \ya in {-.43,0,.43}{%
    \foreach \yb in {-.63,-.21,.21,.63}{%
      \draw[ddgray!85,line width=.27pt] (7.98,\cy+\ya)--(8.49,\cy+\yb);
    }
  }
  \foreach \yb in {-.63,-.21,.21,.63}{%
    \foreach \yc in {-.29,.29}{%
      \draw[ddgray!85,line width=.27pt] (8.49,\cy+\yb)--(9.05,\cy+\yc);
    }
  }
  \foreach \ya in {-.43,0,.43}{\draw[fill=black!3,draw=ddgray,line width=.55pt] (7.98,\cy+\ya) circle (.11);}
  \foreach \yb in {-.63,-.21,.21,.63}{\draw[fill=black!3,draw=ddgray,line width=.55pt] (8.49,\cy+\yb) circle (.11);}
  \foreach \yc in {-.29,.29}{\draw[fill=black!3,draw=ddgray,line width=.55pt] (9.05,\cy+\yc) circle (.11);}
  \draw[->,draw=ddgray,line width=.7pt] (9.24,\cy)--(9.62,\cy);
  \foreach \k/\shade in {0/20,1/55,2/16,3/44,4/68}{%
    \fill[ddblue!\shade!white] (9.74+\k*.24,\cy+.07) rectangle ++(.235,.30);
  }
  \draw[draw=ddblue,line width=.55pt] (9.74,\cy+.07) rectangle (10.94,\cy+.37);
  \foreach \k/\shade in {0/55,1/13,2/29,3/10,4/41}{%
    \fill[ddred!\shade!white] (9.74+\k*.24,\cy-.57) rectangle ++(.235,.30);
  }
  \draw[draw=ddred,line width=.55pt] (9.74,\cy-.57) rectangle (10.94,\cy-.27);
}

\node[note] at (9.37,1.00) {$\ell_2$-normalized features};
\end{scope}

\node[stage] at (14.77,6.82) {Prototype assignments};
\node[stage] at (20.68,6.25) {Joint distribution};

\draw[->,ddgray,line width=.9pt] (11.76,5.08)--(12.96,5.08);
\draw[->,ddgray,line width=.9pt] (11.76,2.29)--(12.96,2.29);

\begin{scope}[xshift=9.2mm]
\foreach \v/\by in {1/4.13,2/1.35}{%
  \draw[rounded corners=.15cm,fill=ddpurplefill,draw=ddpurple!80,
       line width=.70pt] (11.94,\by+.15) rectangle (15.29,\by+1.98);
  \foreach \ra in {.48,1.05,1.61}{%
    \foreach \pr in {.46,.86,1.27,1.67}{%
      \draw[ddpurple!26,line width=.34pt]
        (12.35,\by+\ra)--(14.85,\by+\pr);
    }
  }
  \ifnum\v=1
    \draw[ddpurple!95,line width=1.58pt] (12.35,\by+1.61)--(14.85,\by+1.67);
    \draw[ddpurple!82,line width=1.42pt] (12.35,\by+1.05)--(14.85,\by+.86);
    \draw[ddpurple!86,line width=1.48pt] (12.35,\by+.48)--(14.85,\by+.46);
    \draw[ddpurple!57,line width=.84pt] (12.35,\by+1.05)--(14.85,\by+1.27);
    \draw[ddpurple!52,line width=.76pt] (12.35,\by+.48)--(14.85,\by+.86);
  \else
    \draw[ddpurple!88,line width=1.46pt] (12.35,\by+1.61)--(14.85,\by+1.27);
    \draw[ddpurple!95,line width=1.58pt] (12.35,\by+1.05)--(14.85,\by+1.67);
    \draw[ddpurple!82,line width=1.40pt] (12.35,\by+.48)--(14.85,\by+.86);
    \draw[ddpurple!61,line width=.90pt] (12.35,\by+1.61)--(14.85,\by+1.67);
    \draw[ddpurple!48,line width=.70pt] (12.35,\by+.48)--(14.85,\by+.46);
  \fi
  \foreach \ra in {.48,1.05,1.61}{%
     \draw[fill=ddbluefill,draw=ddblue,line width=.77pt]
          (12.35,\by+\ra) circle (.115);
  }
  \foreach \pr in {.46,.86,1.27,1.67}{%
     \draw[fill=ddredfill,draw=ddred,line width=.78pt,rounded corners=.018cm]
          (14.74,\by+\pr-.105) rectangle (14.95,\by+\pr+.105);
  }
  \node[font=\footnotesize] at (13.62,\by-.20)
       {$Q_{\v}\in\mathbb{R}^{B\times m}$};
}
\end{scope}

\draw[->,ddgray,line width=.90pt] (16.44,5.17) -- (17.69,4.1034);
\draw[->,ddgray,line width=.90pt] (16.44,2.39) -- (17.69,3.5366);
\fill[white] (18.00,3.82) circle (.42);
\draw[draw=ddpurple,line width=.94pt] (18.00,3.82) circle (.42);
\node[font=\large,text=ddpurple!90!black] at (18.00,3.82) {$\times$};
\node[font=\footnotesize,anchor=north] at (18.00,3.25) {$1/B$};
\draw[->,ddgray,line width=.90pt] (18.42,3.82) -- (19.67,3.82);
\node[font=\footnotesize,anchor=south] at (18.945,3.98) {$J$};

\foreach \rowidx in {0,1,2,3}{%
  \foreach \colidx in {0,1,2,3}{%
    \pgfmathtruncatemacro{\sep}{abs(\rowidx+\colidx-3)}
    \pgfmathtruncatemacro{\shade}{max(12,84-29*\sep-2*abs(\rowidx-\colidx))}
    \fill[ddgreen!\shade!white]
       (19.72+\colidx*.48,2.86+\rowidx*.48) rectangle ++(.48,.48);
    \draw[white,line width=.45pt]
       (19.72+\colidx*.48,2.86+\rowidx*.48) rectangle ++(.48,.48);
  }
}
\draw[draw=ddgreen!82,line width=.88pt] (19.72,2.86) rectangle (21.64,4.78);
\node[font=\footnotesize] at (20.68,5.40) {$P=\frac{1}{2}(J+J^\top)$};
\draw[->,ddgray,line width=.90pt] (21.69,3.82) -- (22.94,3.82);
  
\node[textlabel,align=center] (objective) at (24.25,3.82)
  {Maximize\\[3pt]$I_P(C_1;C_2)$};a

\draw[draw=ddred,dashed,line width=1.00pt,->,shorten <=4pt]
  (objective.south)--(24.25,.52)--(1.75,.52)--(1.75,2);
\node[note,text=ddred,fill=white,inner xsep=5pt] at (13.00,.52)
  {Update synthetic pixels};

\end{tikzpicture}%
}
\caption{\textbf{Overview of \textsc{MIPA}.}
Two independent augmentations are applied to both real and synthetic images. A frozen encoder extracts their features to compute assignment distributions of real images over synthetic prototypes, which are used to construct a joint distribution. MIPA maximizes the mutual information between these assignments by updating synthetic images while keeping the encoder fixed.}
\label{fig:mipa-overview}
\label{fig:ssl-dd-pipeline}
\end{figure*}


\subsection{Dataset Distillation}
\label{sec:dd-preliminaries}

\paragraph{Bilevel Formulation.}
Dataset distillation is commonly formulated as a bilevel optimization problem~\cite{wang2018dataset}, where the synthetic data is optimized through its effect on model training. Let $\mathcal{L}^{\mathrm{SSL}}_{U}(\omega)$ denote the expected SSL loss on a dataset $U$, averaged over sampled batches and stochastic augmentations. For $U \in \{\mathcal{S},\mathcal{T}\}$, let $\omega_U = (\theta_U,\phi_U)$ denote the parameters obtained by pretraining on $U$. For a fixed synthetic budget $m$, an SSL-loss formulation is
\begin{equation}
\label{eq:ssldd-bilevel}
\mathcal{S}^{*}
\!\in
\arg\!\min_{\mathcal{S}}
\mathcal{L}^{\mathrm{SSL}}_{\mathcal{T}}(\omega_{\mathcal{S}})
\!\!\quad\text{s.t.}\quad\!\!
\omega_{\mathcal{S}}
\!\in
\arg\!\min_{\omega}
\mathcal{L}^{\mathrm{SSL}}_{\mathcal{S}}(\omega).
\end{equation}
The inner problem trains the encoder and projection head on synthetic data, while the outer problem updates the synthetic examples to reduce the resulting loss on real data.

\paragraph{Representation Matching.}
Although Eq.~\eqref{eq:ssldd-bilevel} does not require labels, its outer objective does not directly compare the representations learned from the synthetic and full datasets. An alternative is to measure the discrepancy between their encoders on real examples~\cite{joshi2025dataset, lee2024self}. Let $f_{\theta_{\mathcal{S}}}$ and $f_{\theta_{\mathcal{T}}}$ denote these encoders. The corresponding objective is
\begin{equation}
\label{eq:ssldd-representation}
\mathcal{S}^{*}
\in
\arg\!\min_{\mathcal{S}}
\frac{1}{|\mathcal{T}|}
\sum_{x \in \mathcal{T}}
d_{\mathrm{rep}}
\left(
f_{\theta_{\mathcal{S}}}(x),
f_{\theta_{\mathcal{T}}}(x)
\right),
\end{equation}
where $d_{\mathrm{rep}}(\cdot,\cdot)$ measures representation discrepancy and $f_{\theta_{\mathcal{T}}}$ is held fixed. This formulation remains bilevel because $\theta_{\mathcal{S}}$ depends on pretraining with the synthetic data.

To evaluate $\mathcal{S}$, the projection head is discarded and the encoder $f_{\theta_{\mathcal{S}}}$ is frozen. 
For each downstream task $k$, the frozen encoder extracts features from the training images. A linear classifier $h_k$ is then trained using these features and the corresponding labels.
The generalization error is
\begin{equation}
\label{eq:downstream-error}
\mathbb{E}_{(x,y)\sim\mathcal{D}_k}
\left[
\mathbf{1}
\left\{
h_k(f_{\theta_{\mathcal{S}}}(x)) \neq y
\right\}
\right],
\end{equation}
where $\mathcal{D}_k$ denotes the distribution of inputs and labels for task $k$, and $\mathbf{1}\{\cdot\}$ is the indicator function. These labels are used only for downstream evaluation. An effective distilled set should achieve a low error across downstream tasks.

\section{Self-Supervised Dataset Distillation}
\label{sec:method}

Directly optimizing Eq.~\eqref{eq:ssldd-representation} is computationally expensive, as it requires repeatedly training an encoder on $\mathcal{S}$. To avoid this inner-loop training, we propose \textbf{M}utual \textbf{I}nformation \textbf{P}rototype \textbf{A}lignment (MIPA). The method uses a frozen encoder pretrained on $\mathcal{T}$ to represent real and synthetic images in the same feature space. The synthetic images serve as \emph{prototypes}, and each augmented real image is assigned to them based on feature similarity. These prototypes are then optimized by maximizing the mutual information between assignments from two augmented views of each real image. 
Figure~\ref{fig:ssl-dd-pipeline} illustrates the complete distillation process.

\subsection{Synthetic Images as Prototypes}
\label{sec:synthetic-prototypes}

Let $f_{\theta_{\mathcal{T}}}$ denote the encoder pretrained on $\mathcal{T}$, whose parameters remain fixed during distillation. For an augmentation $a \sim \mathcal{A}$, its normalized representation is $\psi_a(x) = f_{\theta_{\mathcal{T}}}(a(x))/\|f_{\theta_{\mathcal{T}}}(a(x))\|_2$. Each synthetic image $x_j^s$ defines a prototype through $\psi_a(x_j^s)$. A real image $x$ is assigned to these prototypes according to cosine similarity,
\begin{equation}
\label{eq:prototype-assignment}
q(j\mid x,\mathcal{S},a)
=
\frac{
\exp\!\left(\psi_a(x)^\top\psi_a(x_j^s)/\tau\right)
}{
\sum_{r=1}^{m}
\exp\!\left(\psi_a(x)^\top\psi_a(x_r^s)/\tau\right)
},
\end{equation}
where $\tau>0$ controls the sharpness of the assignments. Collecting these probabilities gives
\begin{equation}
\label{eq:surrogate_rep}
\mathbf{q}_{\mathcal{S}}(x,a)
=
\left[
q(1\mid x,\mathcal{S},a),
\ldots,
q(m\mid x,\mathcal{S},a)
\right]^\top.
\end{equation}
This vector represents a real image through its similarity to the synthetic set. Unlike $f_{\theta_{\mathcal{S}}}(x)$, it depends directly on the synthetic images without requiring an encoder to be trained on them.

For each real batch $\mathcal{B}$, two transformations $a^{(1)},a^{(2)} \sim \mathcal{A}$ are sampled independently. Within each view, the same transformation and its sampled parameters are applied to all real and synthetic images. This compares images and prototypes under matched augmentations. In particular, when $x_j^s=x$, their normalized features coincide and their cosine similarity is one. Independent transformations can reduce this similarity even for identical images.

A useful prototype representation should retain distinctions between real images while remaining consistent across augmented views. Let $\bar{\mathbf{q}}_{\mathcal{S}}(x) = \mathbb{E}_{a\sim\mathcal{A}}[\mathbf{q}_{\mathcal{S}}(x,a)]$ denote the mean assignment of $x$. Consistency across augmentations can be measured by
\begin{equation}
\label{eq:rep_variance}
\mathcal{V}(\mathcal{S})
=
\mathbb{E}_{x\sim\mathcal{T},\,a\sim\mathcal{A}}
\left[
\left\|
\mathbf{q}_{\mathcal{S}}(x,a)
-
\bar{\mathbf{q}}_{\mathcal{S}}(x)
\right\|_2^2
\right],
\end{equation}
where $x\sim\mathcal{T}$ denotes uniform sampling from the real dataset. For two independently sampled augmentations, the expected squared difference between their assignments equals twice this variance. For a batch of $B$ real images,
\begin{equation}
\label{eq:l_agree}
\mathcal{L}_{\mathrm{agree}}
=
\frac{1}{2B}
\sum_{x\in\mathcal{B}}
\left\|
\mathbf{q}_{\mathcal{S}}(x,a^{(1)})
-
\mathbf{q}_{\mathcal{S}}(x,a^{(2)})
\right\|_2^2
\end{equation}
estimates $\mathcal{V}(\mathcal{S})$.
However, agreement alone admits trivial solutions. Assigning every image to the same constant vector makes $\mathcal{L}_{\mathrm{agree}}$ zero without preserving distinctions between images. 
Therefore, assignment consistency alone is insufficient and does not constitute an equivalent reformulation of encoder matching.

\subsection{Mutual Information Maximization}
\label{sec:information-objective}

To encourage both cross-view predictability and diverse prototype usage, we maximize the mutual information between paired assignments. This follows the information-maximization principle used in~\cite{9008122} for unsupervised clustering.
For each view $v$, let $\mathbf{Q}_v \in \mathbb{R}^{B\times m}$ contain the assignment vectors as rows, with $[\mathbf{Q}_v]_{ij}=q(j\mid x_i,\mathcal{S},a^{(v)})$. The empirical joint probability matrix is
\begin{equation}
\label{eq:joint-assignment}
J
=
\frac{1}{B}\mathbf{Q}_1^\top\mathbf{Q}_2,
\quad
P
=
\frac{1}{2}(J+J^\top).
\end{equation}
Each entry $J_{jr}$ averages the product of assignment probabilities to prototype $j$ in the first view and prototype $r$ in the second. Symmetrizing $J$ treats both view orders equally and gives identical row and column marginals, $p_j=\sum_{r=1}^{m}P_{jr}=\sum_{r=1}^{m}P_{rj}$.

Let $C_1,C_2\in\{1,\ldots,m\}$ denote prototype indices with joint distribution $P$. Their mutual information is
\begin{equation}
\label{eq:mutual-information}
I_P(C_1;C_2)
=
\sum_{j,r=1}^{m}
P_{jr}
\log
\frac{P_{jr}}{p_jp_r}.
\end{equation}
Using $I_P(C_1;C_2)=H_P(C_1)-H_P(C_1\mid C_2)$, where $H_P$ denotes entropy under $P$, the objective has two complementary effects. Lower conditional entropy makes the assignment in one view more predictable from the other while higher marginal entropy encourages the use of multiple prototypes. A constant assignment shared by all images and both views gives zero mutual information.
\vspace{-0.6em}

\paragraph{Distillation Objective.}
The synthetic images determine the prototype features and consequently the assignments in Eq.~\eqref{eq:prototype-assignment}. We optimize these images to maximize mutual information between paired assignments. Gradients update the synthetic pixels through the prototype features, without training an encoder on $\mathcal{S}$. The resulting objective is
\begin{equation}
\label{eq:cross-view-distillation}
\mathcal{S}^{*}
\in
\arg\!\max_{\mathcal{S}}
\mathbb{E}_{\mathcal{B},\,a^{(1)},\,a^{(2)}}
\left[
I_P(C_1;C_2)
\right],
\end{equation}
where the expectation is over real batches $\mathcal{B}$ and independently sampled augmentations $a^{(1)},a^{(2)}\sim\mathcal{A}$. For each batch and augmentation pair, $P$ is computed from the prototype assignments using Eq.~\eqref{eq:joint-assignment}.

\newcommand{\mipatable}{%
  \footnotesize
  \renewcommand{\arraystretch}{1.05}%
}
\newcommand{\mipaacc}[2]{#1{\scriptsize$\pm$#2}}
\newcommand{\mipabest}[2]{\textbf{#1}{\scriptsize$\pm$#2}}

\section{Experiments}
\subsection{Experimental Settings}

\begin{table*}[t]
\centering
\caption{\textbf{Size (\%)} denotes the fraction of the full training set used for pretraining. \textbf{Labeled Data} denotes the percentage of labeled data available for training the linear probe. \textbf{\textsc{Full Data}} refers to SSL pretraining on the full training set and serves as a performance ceiling. Results are reported as the mean $\pm$ standard deviation across five runs.
Best results among reduced-data methods are in \textbf{bold}.
}
\vspace{-0.6em}
\label{tab:resnet18-main-all}
\mipatable
\newcommand{\mipamethod}[1]{\hspace*{-1pt}#1\hspace*{1pt}}
\renewcommand{\mipaacc}[2]{\hspace*{0.6pt}#1\,\textcolor{gray!90}{\fontsize{7.5}{6.2}\selectfont$\pm$#2}\hspace*{-0.6pt}}
\renewcommand{\mipabest}[2]{\hspace*{0.6pt}\textbf{#1}\,\textcolor{gray!90}{\fontsize{7.5}{6.2}\selectfont$\pm$#2}\hspace*{-0.6pt}}
\setlength{\tabcolsep}{3pt}
\renewcommand{\arraystretch}{1.2}

\resizebox{\textwidth}{!}{%
\begin{tabular}{@{}c@{\hspace{5pt}}c@{\hspace{5pt}}*{7}{r}@{\hspace{7pt}}*{7}{r}@{}}
\toprule
\multirow{2}{*}[-2.5pt]{Dataset}
& \multirow{2}{*}[-2.5pt]{\shortstack[c]{Size\\(\%)}}
& \multicolumn{7}{c}{{1\% labeled data}}
& \multicolumn{7}{c}{{5\% labeled data}} \\
\cmidrule(lr){3-9}\cmidrule(l){10-16}
& & \multicolumn{1}{c}{\mipamethod{\textsc{Random}}}
& \multicolumn{1}{c}{\mipamethod{\textsc{K-means}}}
& \multicolumn{1}{c}{\mipamethod{\textsc{SAS}}}
& \multicolumn{1}{c}{\mipamethod{\textsc{KRR-ST}}}
& \multicolumn{1}{c}{\mipamethod{\textsc{MKDT}}}
& \multicolumn{1}{c}{\mipamethod{\textbf{\textsc{MIPA}}}}
& \multicolumn{1}{c@{\hspace{7pt}}}{\mipamethod{\textsc{Full Data}}}
& \multicolumn{1}{c}{\mipamethod{\textsc{Random}}}
& \multicolumn{1}{c}{\mipamethod{\textsc{K-means}}}
& \multicolumn{1}{c}{\mipamethod{\textsc{SAS}}}
& \multicolumn{1}{c}{\mipamethod{\textsc{KRR-ST}}}
& \multicolumn{1}{c}{\mipamethod{\textsc{MKDT}}}
& \multicolumn{1}{c}{\mipamethod{\textbf{\textsc{MIPA}}}}
& \multicolumn{1}{c@{}}{\mipamethod{\textsc{Full Data}}} \\
\midrule

\multirow{3}{*}{CIFAR-10} & 0.2
& \mipaacc{40.04}{0.45}
& \mipaacc{38.77}{0.95}
& \mipaacc{37.91}{0.79}
& \mipaacc{28.22}{0.92}
& \mipaacc{40.12}{0.67}
& \mipabest{45.18}{0.41}
& \multirow{3}{*}{\mipaacc{87.23}{0.69}}
& \mipaacc{46.20}{0.27}
& \mipaacc{45.19}{0.41}
& \mipaacc{44.39}{0.63}
& \mipaacc{34.23}{0.49}
& \mipaacc{46.76}{0.40}
& \mipabest{50.58}{0.47}
& \multirow{3}{*}{\mipaacc{89.03}{0.26}} \\
& 1
& \mipaacc{48.34}{0.63}
& \mipaacc{46.10}{0.68}
& \mipaacc{44.56}{0.44}
& \mipaacc{29.07}{0.78}
& \mipaacc{40.29}{0.99}
& \mipabest{52.26}{0.55}
&
& \mipaacc{52.05}{1.00}
& \mipaacc{51.80}{0.93}
& \mipaacc{52.17}{0.30}
& \mipaacc{35.93}{0.60}
& \mipaacc{46.66}{0.67}
& \mipabest{57.36}{0.16}
& \\
& 2
& \mipaacc{54.10}{0.38}
& \mipaacc{52.98}{0.57}
& \mipaacc{49.87}{0.81}
& \mipaacc{41.26}{0.55}
& \mipaacc{40.56}{0.92}
& \mipabest{59.31}{0.40}
&
& \mipaacc{58.24}{0.12}
& \mipaacc{57.77}{0.37}
& \mipaacc{55.49}{0.26}
& \mipaacc{47.27}{0.59}
& \mipaacc{46.55}{0.86}
& \mipabest{62.53}{0.15}
& \\
\midrule

\multirow{3}{*}{CIFAR-100} & 0.2
& \mipaacc{9.91}{0.54}
& \mipaacc{8.51}{0.65}
& \mipaacc{10.41}{0.57}
& \mipaacc{8.28}{0.55}
& \mipaacc{9.43}{0.81}
& \mipabest{10.65}{0.46}
& \multirow{3}{*}{\mipaacc{48.12}{0.86}}
& \mipaacc{16.87}{0.33}
& \mipaacc{15.25}{0.24}
& \mipaacc{17.12}{0.08}
& \mipaacc{14.75}{0.46}
& \mipaacc{16.11}{0.31}
& \mipabest{17.17}{0.07}
& \multirow{3}{*}{\mipaacc{57.01}{0.41}} \\
& 1
& \mipaacc{12.45}{0.25}
& \mipaacc{11.79}{0.48}
& \mipaacc{13.29}{0.70}
& \mipaacc{9.62}{0.51}
& \mipaacc{10.72}{0.61}
& \mipabest{13.76}{0.39}
&
& \mipaacc{20.50}{0.24}
& \mipaacc{19.50}{0.34}
& \mipaacc{20.39}{0.28}
& \mipaacc{16.89}{0.42}
& \mipaacc{18.20}{0.65}
& \mipabest{21.70}{0.36}
& \\
& 2
& \mipaacc{14.42}{0.95}
& \mipaacc{15.00}{0.96}
& \mipaacc{16.01}{1.02}
& \mipaacc{14.11}{0.69}
& \mipaacc{7.53}{0.39}
& \mipabest{19.15}{0.78}
&
& \mipaacc{23.15}{0.47}
& \mipaacc{24.07}{0.31}
& \mipaacc{24.68}{0.21}
& \mipaacc{22.17}{0.21}
& \mipaacc{14.14}{0.57}
& \mipabest{27.58}{0.18}
& \\
\midrule

\multirow{3}{*}{\shortstack[c]{Tiny\\ImageNet}} & 0.2
& \mipaacc{5.46}{0.20}
& \mipaacc{5.00}{0.30}
& \mipaacc{5.42}{0.22}
& \mipaacc{5.47}{0.23}
& \mipaacc{5.33}{0.26}
& \mipabest{5.98}{0.27}
& \multirow{3}{*}{\mipaacc{28.78}{0.57}}
& \mipaacc{9.88}{0.07}
& \mipaacc{9.23}{0.27}
& \mipaacc{9.54}{0.22}
& \mipaacc{10.30}{0.33}
& \mipaacc{10.14}{0.31}
& \mipabest{10.49}{0.30}
& \multirow{3}{*}{\mipaacc{39.81}{0.68}} \\
& 1
& \mipaacc{7.18}{0.26}
& \mipaacc{6.62}{0.20}
& \mipaacc{7.23}{0.30}
& \mipaacc{6.45}{0.19}
& \mipaacc{5.76}{0.20}
& \mipabest{8.63}{0.17}
&
& \mipaacc{13.90}{0.18}
& \mipaacc{13.49}{0.28}
& \mipaacc{13.42}{0.20}
& \mipaacc{12.78}{0.44}
& \mipaacc{11.20}{0.15}
& \mipabest{15.11}{0.20}
& \\
& 2
& \mipaacc{10.87}{0.37}
& \mipaacc{10.37}{0.26}
& \mipaacc{10.85}{0.27}
& \mipaacc{9.85}{0.32}
& \mipaacc{5.73}{0.16}
& \mipabest{11.77}{0.06}
&
& \mipaacc{18.21}{0.21}
& \mipaacc{17.91}{0.30}
& \mipaacc{18.08}{0.13}
& \mipaacc{16.97}{0.32}
& \mipaacc{11.63}{0.16}
& \mipabest{19.10}{0.22}
& \\

\bottomrule
\end{tabular}%
}
\end{table*}

\paragraph{Datasets and baselines.}
We evaluate \textsc{MIPA} on CIFAR-10, CIFAR-100~\cite{krizhevsky2009learning}, and Tiny ImageNet~\cite{le2015tiny}, and conduct downstream evaluations on these datasets as well as FGVC Aircraft~\cite{maji2013fine} and Stanford Dogs~\cite{khosla2011novel}. 
We compare against three sample selection methods, \textsc{Random}, \textsc{K-means}, and \textsc{SAS}~\cite{joshi2023data}, as well as two SSL-DD methods \textsc{KRR-ST}~\cite{lee2024self} and  \textsc{MKDT}~\cite{joshi2025dataset}. \textsc{Random} selects real training images uniformly at random, whereas \textsc{K-means} performs clustering over pretrained encoder embeddings of the full training set and then chooses real images closest to the resulting centroids.
\textsc{SAS} also selects real samples based on their expected similarity to augmented views of other training examples, prioritizing those that are most beneficial for contrastive SSL. For \textsc{KRR-ST} and \textsc{MKDT}, 
we generate distilled sets using their official implementations.


For a consistent image-only comparison, we evaluate \textsc{KRR-ST} and \textsc{MKDT} using their synthetic images alone, excluding the auxiliary target representations required by their original procedures. This evaluates the standalone utility of the distilled images under a common from-scratch SSL protocol, rather than the performance of their complete training pipelines. We focus on this setting because the resulting datasets can be reused with standard SSL objectives without method-specific auxiliary targets. Methods that train a classifier on top of a frozen pretrained encoder (e.g., \cite{zhou2024scdd,cazenavette2025pretrained}) address a different downstream setting and are not directly comparable here. We also report SSL results using the full training set as an upper-bound reference.
\vspace{-2em}

\paragraph{Distillation setup.}
We first pretrain a ResNet-18~\cite{he2016deep} encoder with BatchNorm on the full unlabeled training set using SimCLR~\cite{chen2020simple} for $1k$ epochs and keep it frozen for the whole distillation process. Prototype assignments are then obtained from cosine similarities between embeddings produced by this frozen teacher from augmented views of real and synthetic images using a temperature $\tau$ of $0.2$.
Unless otherwise noted, synthetic data are initialized by randomly sampling real images from the full dataset, and then optimized with Adam using a learning rate of $0.001$ and a batch size $\mathcal{B}$ of $128$.
We consider synthetic dataset sizes of $0.2\%$, $1\%$, and $2\%$ of the full training set and run \textsc{MIPA} for $10k$, $8k$, and $6k$ iterations, respectively.
\vspace{-0.6em}


\paragraph{Evaluation setup.}
For each method and dataset budget, we train a ResNet-18 encoder from scratch using SimCLR on the resulting images alone. For \textsc{KRR-ST} and \textsc{MKDT}, this excludes the additional target representations required by their original downstream SSL procedures. We assess classification on the source dataset and transfer to downstream tasks through linear probing. In each case, we freeze the encoder and train a linear probe using either $1\%$ or $5\%$ from the labeled training data of the evaluation dataset, then measure accuracy on its test set. We repeat each evaluation over five independent runs and report the mean accuracy.

\subsection{Results and Discussion}

\paragraph{Main results.}
Table~\ref{tab:resnet18-main-all} shows that \textsc{MIPA} achieves the highest mean accuracy among reduced-data methods across all reported configurations, although some improvements over the strongest baselines are small relative to the reported variability.
Its improvements over the sample selection baselines suggest the benefit of optimizing synthetic images rather than selecting real samples. The gains are consistent on CIFAR-10 and become more pronounced on CIFAR-100 as the synthetic set size increases. At the $2\%$ CIFAR-100 setting, \textsc{MIPA} exceeds the strongest baseline, \textsc{SAS}, by $3.14$ and $2.90$ percentage points with $1\%$ and $5\%$ labeled data, respectively. Gains on Tiny ImageNet are smaller but remain consistent across the evaluated settings.

\textsc{MIPA} also consistently benefits from larger synthetic datasets, whereas \textsc{MKDT} shows only marginal improvements or even declines in accuracy as the dataset size increases.
One possible explanation is that \textsc{MKDT} optimizes synthetic images to be used alongside teacher-derived representations, so increasing the synthetic budget does not necessarily improve their standalone utility when these targets are removed.
Both \textsc{KRR-ST} and \textsc{MKDT} generally underperform \textsc{Random} under our image-only evaluation protocol, with \textsc{KRR-ST} showing particularly large performance gaps on CIFAR-10. Importantly, both methods originally rely on auxiliary target representations alongside synthetic images for pretraining. Their lower performance in our setting therefore reflects the limitations of using their distilled images alone, rather than their complete training procedures. Overall, \textsc{MIPA} outperforms these methods using only synthetic images for SSL pretraining from scratch, although a gap to full-data pretraining still remains.

\begin{figure*}[t]
\centering
\definecolor{downrandom}{HTML}{4D82AC}
\definecolor{downkmeans}{HTML}{568F7D}
\definecolor{downmkdt}{HTML}{8E78A0}
\definecolor{downmipa}{HTML}{C45E3C}
\begin{minipage}{\textwidth}
  \centering
  \includegraphics[width=1.65in]{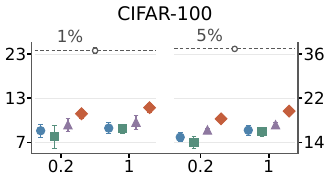}\hfill%
  \includegraphics[width=1.65in]{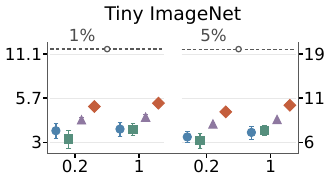}\hfill%
  \includegraphics[width=1.65in]{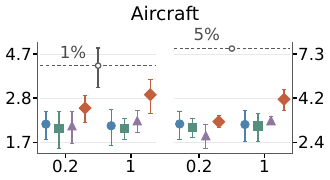}\hfill%
  \includegraphics[width=1.65in]{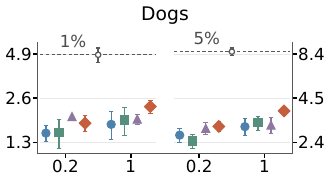}
  \captionsetup{font=footnotesize,skip=2pt}
  \caption*{CIFAR-10 pretraining}
\end{minipage}
\par\vspace{4bp}
\begin{minipage}{\textwidth}
  \centering
  \includegraphics[width=1.65in]{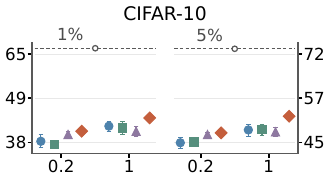}\hfill%
  \includegraphics[width=1.65in]{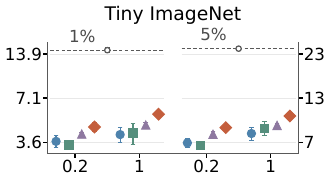}\hfill%
  \includegraphics[width=1.65in]{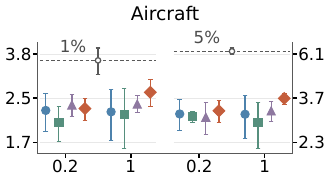}\hfill%
  \includegraphics[width=1.65in]{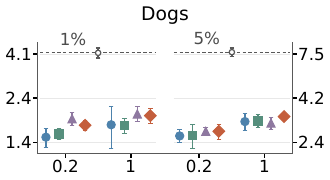}
  \captionsetup{font=footnotesize,skip=2pt}
  \caption*{CIFAR-100 pretraining}
\end{minipage}
\par\vspace{4bp}
\begin{minipage}{\textwidth}
  \centering
  \includegraphics[width=1.65in]{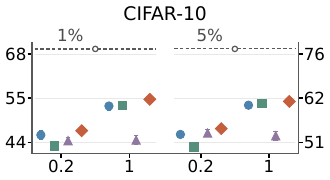}\hfill%
  \includegraphics[width=1.65in]{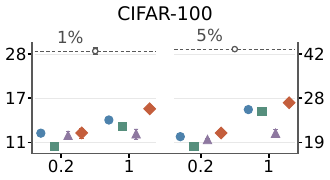}\hfill%
  \includegraphics[width=1.65in]{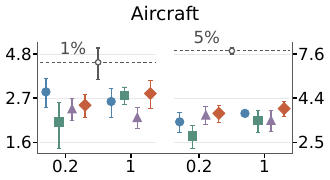}\hfill%
  \includegraphics[width=1.65in]{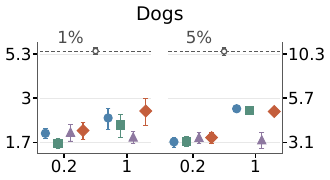}
  \captionsetup{font=footnotesize,skip=2pt}
  \caption*{Tiny ImageNet pretraining}
\end{minipage}
\caption{\textbf{Cross-dataset transfer.} Titles show downstream tasks; rows identify pretraining data. Ticks 0.2 and 1 denote pretraining percentages. The 5\% and 1\% downstream label budgets use independent left and right accuracy (\%) axes. \textcolor{downrandom}{$\bullet$} \textsc{Random}; \textcolor{downkmeans}{$\blacksquare$} \textsc{K-means}; \textcolor{downmkdt}{$\blacktriangle$} \textsc{MKDT}; \textcolor{downmipa}{$\blacklozenge$} \textsc{MIPA}. Dashed gray lines with open circles show \textsc{Full Data} pretraining; whiskers show sample standard deviation.}
\label{fig:downstream-transfer}
\end{figure*}
\vspace{-1em}
\paragraph{Cross-dataset generalization.}
In this section, we evaluate how well models trained on synthetic datasets distilled by \textsc{MIPA} generalize when the downstream evaluation dataset differs from that used for SSL pretraining. Figure~\ref{fig:downstream-transfer} illustrates the results.
As shown in the figure, \textsc{MIPA} matches or outperforms the other evaluated methods in every transfer setting among CIFAR-10, CIFAR-100, and Tiny ImageNet across different configurations.
Increasing the synthetic set size also consistently improves \textsc{MIPA}'s downstream accuracy across all evaluated settings, indicating that the advantages of a larger distilled set generalize beyond classification on the source dataset.
Transfer to Aircraft and Dogs, however, is less stable, as \textsc{MKDT} or \textsc{Random} can outperform \textsc{MIPA} in some settings.
Overall, these results indicate that \textsc{MIPA} continues to perform well in downstream evaluations, although performance gaps between methods are often relatively minor. 
\vspace{-1em}
\paragraph{Cross-architecture generalization.}
We evaluate whether the gains of \textsc{MIPA} persist when pretraining deeper networks than the one used to generate synthetic data. 
We use ResNet-18 as the frozen encoder during distillation and the resulting synthetic data to pretrain ResNet-34 and ResNet-50 encoders.
Table~\ref{tab:cross-arch-combined} reports the results.
\textsc{MIPA} outperforms \textsc{Random} on the source dataset in every evaluated setting, showing its gains extend to encoders different from the one used during distillation. Downstream accuracy also improves across all ResNet-50 and most ResNet-34 settings. 
With ResNet-34, the larger CIFAR-10 synthetic set yields higher source-dataset accuracy but slightly lower downstream accuracy on CIFAR-100 and Tiny ImageNet. \textsc{MIPA} also performs slightly worse than \textsc{Random} when transferring from the smaller CIFAR-100 synthetic data to CIFAR-10.
Overall, \textsc{MIPA} maintains its benefits across architectures, with a few exceptions in downstream tasks.

\vspace{-0.4em}
\begin{figure}[h!]
    \centering
    \includegraphics[width=0.49\columnwidth]
        {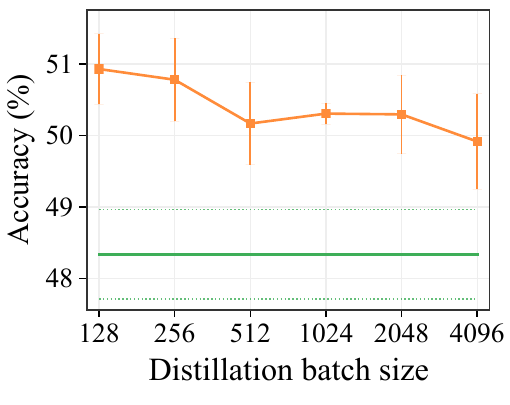}
    \hfill
    \includegraphics[width=0.49\columnwidth]
        {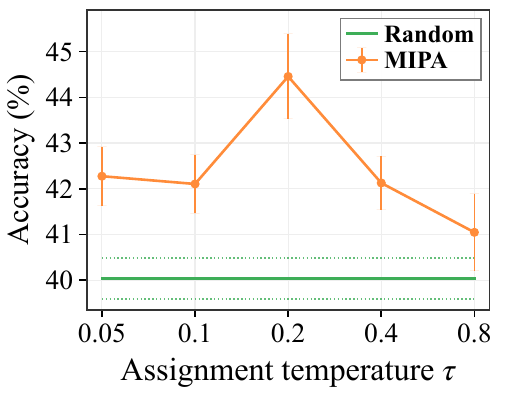}
\vspace{-0.6em}
\caption{Effect of batch size (left, $1\%$ data) and assignment temperature (right, $0.2\%$ data) on CIFAR-10. Curves show mean accuracy $\pm$ SD.}
    \label{fig:mipa-batch-temperature}
\end{figure}
\vspace{-1.5em}
\paragraph{Effect of batch size and assignment temperature.}
We analyze how batch size and assignment temperature $\tau$ in Eqs.~\eqref{eq:prototype-assignment} and~\eqref{eq:joint-assignment} affect performance. 
On CIFAR-10, we vary batch size using a $1\%$ synthetic set and assignment temperature using a $0.2\%$ synthetic set, keeping other settings fixed.
Figure~\ref{fig:mipa-batch-temperature} shows that \textsc{MIPA} outperforms \textsc{Random} in all evaluated settings.
However, increasing batch size does not necessarily improve performance. This may occur because averaging the joint assignment matrix $P$ over large batches smooths the empirical marginal distribution, weakening gradient signals from the marginal entropy term $H_P(C_1)$ that encourages diverse prototype usage.
Turning to the assignment temperature, the sweep shows that an intermediate value of $\tau=0.2$ performs best. Smaller temperatures produce overly sharp assignments, while larger temperatures make assignments more uniform across synthetic images. An intermediate temperature may therefore preserve distinctions between real images while avoiding excessive concentration on individual prototypes.
\vspace{-0.6em}

\begin{table*}[t]
\centering
\caption{Cross-architecture generalization.
Synthetic data are generated using ResNet-18 teachers and evaluated on deeper networks. C10, C100, and TIN denote CIFAR-10, CIFAR-100, and Tiny ImageNet, respectively.}
\vspace{-0.6em}
\label{tab:cross-arch-combined}

\normalsize
\setlength{\tabcolsep}{1.5pt}
\renewcommand{\arraystretch}{1.02}

\def\backbonegap{10pt}     
\def\methodgap{3pt}        
\def\resultgap{8pt}        
\def\datasetgap{0pt}      
\def\labelgap{5pt}         
\def\pretraingap{10pt}     

\def\acc#1#2{#1\,\textcolor{gray!90}{\scriptsize$\pm$#2}}
\def\best#1#2{\textbf{#1}\,\textcolor{gray!90}{\scriptsize$\pm$#2}}

\resizebox{\textwidth}{!}{%
\begin{tabular}{@{}
l!{\hspace{\backbonegap}}
c!{\hspace{\methodgap}}
l!{\hspace{\resultgap}}
c!{\hspace{\datasetgap}}c!{\hspace{\datasetgap}}c
!{\hspace{\labelgap}}
c!{\hspace{\datasetgap}}c!{\hspace{\datasetgap}}c
!{\hspace{\pretraingap}}
c!{\hspace{\datasetgap}}c!{\hspace{\datasetgap}}c
!{\hspace{\labelgap}}
c!{\hspace{\datasetgap}}c!{\hspace{\datasetgap}}c@{}}

\toprule
\multirow{3}{*}{} &
\multirow{3}{*}{\vspace{-1.3em}\shortstack{Size \\{(\%)}}} &
\multirow{3}{*}{Method} &
\multicolumn{6}{c}{{CIFAR-10 Pretraining}} &
\multicolumn{6}{c}{{CIFAR-100 Pretraining}} \\
\cmidrule(lr){4-9}\cmidrule(l){10-15}

& & &
\multicolumn{3}{c}{1\% Labels} &
\multicolumn{3}{c}{5\% Labels} &
\multicolumn{3}{c}{1\% Labels} &
\multicolumn{3}{c}{5\% Labels} \\
\cmidrule(lr){4-6}\cmidrule(lr){7-9}
\cmidrule(lr){10-12}\cmidrule(l){13-15}

& & &
C10 & C100 & TIN &
C10 & C100 & TIN &
C10 & C100 & TIN &
C10 & C100 & TIN \\
\midrule

\multirow{5}{*}{\vspace{-0.5em}\rotatebox[origin=c]{90}{\large{ResNet-34}}}
& 100 & \textsc{Full Data}
& \acc{88.66}{0.20} & \acc{24.06}{0.70} & \acc{12.14}{0.24}
& \acc{90.39}{0.15} & \acc{37.95}{0.22} & \acc{20.73}{0.20}
& \acc{66.70}{0.26} & \acc{49.60}{0.56} & \acc{14.95}{0.19}
& \acc{74.22}{0.31} & \acc{58.09}{0.39} & \acc{24.57}{0.28} \\

\cmidrule(l){2-15}

& \multirow{2}{*}{0.2} & \textsc{Random}
& \acc{39.39}{1.77} & \acc{9.27}{0.57} & \acc{3.99}{0.29}
& \acc{46.49}{1.38} & \acc{16.02}{0.44} & \acc{7.71}{0.23}
& \best{39.12}{0.67} & \acc{10.15}{0.36} & \acc{4.00}{0.13}
& \acc{45.77}{0.43} & \acc{16.38}{0.38} & \acc{7.69}{0.21} \\

& & \textbf{MIPA}
& \best{43.15}{0.56} & \best{9.84}{0.42} & \best{4.43}{0.15}
& \best{49.18}{0.39} & \best{16.94}{0.11} & \best{8.35}{0.10}
& \acc{38.88}{0.58} & \best{11.14}{0.29} & \best{4.19}{0.13}
& \best{46.18}{0.50} & \best{16.99}{0.24} & \best{7.92}{0.29} \\ 
\arrayrulecolor{black!65}
\cmidrule[0.1pt](l){3-15}
\arrayrulecolor{black}
& \multirow{2}{*}{1} & \textsc{Random}
& \acc{47.20}{0.25} & \best{11.39}{0.29} & \acc{4.71}{0.13}
& \acc{53.54}{0.17} & \best{18.59}{0.39} & \best{9.15}{0.26}
& \acc{41.98}{0.43} & \acc{12.48}{0.46} & \acc{4.92}{0.17}
& \acc{49.52}{0.24} & \acc{19.37}{0.24} & \acc{9.21}{0.24} \\

& & \textbf{MIPA}
& \best{51.61}{0.56} & \acc{10.86}{0.32} & \best{4.91}{0.25}
& \best{56.95}{0.20} & \acc{18.55}{0.15} & \acc{8.96}{0.23}
& \best{42.64}{0.61} & \best{14.56}{0.14} & \best{5.30}{0.13}
& \best{50.11}{0.17} & \best{21.32}{0.52} & \best{9.61}{0.35} \\

\midrule

\multirow{5}{*}{\vspace{-0.5em}\rotatebox[origin=c]{90}{\large{ResNet-50}}}
& 100 & \textsc{Full Data}
& \acc{87.78}{0.26} & \acc{26.86}{1.07} & \acc{12.70}{0.26}
& \acc{90.28}{0.18} & \acc{42.35}{0.15} & \acc{22.83}{0.23}
& \acc{69.07}{0.32} & \acc{49.66}{0.59} & \acc{15.09}{0.31}
& \acc{77.60}{0.16} & \acc{59.63}{0.40} & \acc{25.65}{0.28} \\

\cmidrule(l){2-15}

& \multirow{2}{*}{0.2} & \textsc{Random}
& \acc{37.48}{1.11} & \acc{8.43}{0.55} & \acc{3.95}{0.18}
& \acc{44.20}{1.24} & \acc{15.33}{0.12} & \acc{8.07}{0.11}
& \acc{36.68}{0.75} & \acc{9.22}{0.55} & \acc{4.06}{0.24}
& \acc{43.76}{0.43} & \acc{15.59}{0.51} & \acc{8.09}{0.21} \\

& & \textbf{MIPA}
& \best{41.09}{0.72} & \best{9.72}{0.83} & \best{4.70}{0.32}
& \best{47.04}{0.34} & \best{16.86}{0.13} & \best{9.18}{0.15}
& \best{38.84}{0.54} & \best{10.38}{0.26} & \best{4.84}{0.21}
& \best{46.63}{0.45} & \best{17.21}{0.47} & \best{8.90}{0.17} \\ 

\arrayrulecolor{black!65}
\cmidrule[0.1pt](l){3-15}
\arrayrulecolor{black}

& \multirow{2}{*}{1} & \textsc{Random}
& \acc{46.09}{0.56} & \acc{11.78}{0.29} & \acc{5.13}{0.19}
& \acc{54.00}{0.43} & \acc{19.96}{0.34} & \acc{10.26}{0.21}
& \acc{42.66}{0.28} & \acc{12.47}{0.58} & \acc{5.30}{0.24}
& \acc{51.88}{0.43} & \acc{20.82}{0.44} & \acc{10.25}{0.41} \\

& & \textbf{MIPA}
& \best{49.85}{0.29} & \best{12.13}{0.63} & \best{5.42}{0.15}
& \best{57.11}{0.26} & \best{20.62}{0.26} & \best{10.55}{0.23}
& \best{43.56}{0.47} & \best{13.48}{0.47} & \best{5.58}{0.14}
& \best{52.47}{0.71} & \best{22.40}{0.66} & \best{10.69}{0.34} \\

\bottomrule
\end{tabular}
}
\end{table*}

\vspace{-0.3em}
\paragraph{Ablation on the distillation objective.}
In Section~\ref{sec:synthetic-prototypes}, we explained that naively enforcing assignment agreement between augmented views can lead to collapse and trivial solutions. Here, we empirically verify this behavior. Specifically, we perform distillation using the agreement objective $\mathcal{L}_{\mathrm{agree}}$ from Eq.~\eqref{eq:l_agree} and compare it with \textsc{Random} and \textsc{MIPA}. Table~\ref{tab:mipa-components} summarizes the results, indicating that optimizing $\mathcal{L}_{\mathrm{agree}}$ alone causes collapse and yields performance worse than \textsc{Random}. This finding is consistent with our analysis and supports the use of mutual information to promote diverse prototype assignments while maintaining agreement between augmented views.

\vspace{-0.6em}
\begin{table}[!h]
\centering
\caption{Distillation objective ablation on CIFAR-10 with ResNet-18 and SimCLR. Results are linear probing accuracy (\%, mean $\pm$ std.\ over five runs). Bold marks the best result.}
\vspace{-0.6em}
\label{tab:mipa-components}
\mipatable
\setlength{\tabcolsep}{1.1pt}
\renewcommand{\arraystretch}{1}
\begin{tabular*}{\columnwidth}{@{\extracolsep{\fill}}l*{4}{r}@{}}
\toprule
\multirow{2}{*}[-2.5pt]{Method} & \multicolumn{2}{c}{\textbf{0.2\% data}} & \multicolumn{2}{c}{\textbf{1\% data}} \\
\cmidrule(lr){2-3}\cmidrule(l){4-5}
 & 1\% labels & 5\% labels & 1\% labels & 5\% labels \\
\midrule
\textsc{Random} & \mipaacc{40.04}{0.45} & \mipaacc{46.20}{0.27} & \mipaacc{48.34}{0.63} & \mipaacc{52.05}{1.00} \\
Agreement $\mathcal{L}_{\mathrm{agree}}$ & \mipaacc{37.60}{0.82} & \mipaacc{43.53}{0.62} & \mipaacc{40.69}{0.82} & \mipaacc{47.65}{0.21} \\
{MIPA} & \mipabest{45.18}{0.41} & \mipabest{50.58}{0.47} & \mipabest{52.26}{0.55} & \mipabest{57.36}{0.16} \\
\bottomrule
\end{tabular*}
\end{table}

\vspace{-1em}
\paragraph{Evaluation with Barlow Twins.}
Next, we evaluate whether the benefits of \textsc{MIPA} generalize to an SSL method beyond SimCLR. Specifically, we pretrain an encoder from scratch 
using Barlow Twins on synthetic data produced by \textsc{MIPA}, where the frozen teacher was trained with SimCLR.
Table~\ref{tab:ssl-methods-combined} presents the results, indicating that \textsc{MIPA} consistently surpasses \textsc{Random} on the source dataset across all settings and in every downstream evaluation except one. It also consistently outperforms \textsc{MKDT}.
These results indicate that images distilled using a teacher trained with SimCLR remain effective for pretraining with Barlow Twins.

\begin{table*}[!t]
\centering
\caption{Ablation study on generalization to Barlow Twins SSL. C10, C100, and TIN denote CIFAR-10, CIFAR-100, and Tiny ImageNet.}
\label{tab:ssl-methods-combined}
\vspace{-0.6em}

\normalsize
\setlength{\tabcolsep}{1.5pt}
\renewcommand{\arraystretch}{1}

\def\methodgap{5pt}        
\def\resultgap{8pt}        
\def\datasetgap{0pt}       
\def\labelgap{5pt}         
\def\pretraingap{10pt}     

\def\acc#1#2{#1\,\textcolor{gray!90}{\scriptsize$\pm$#2}}
\def\best#1#2{\textbf{#1}\,\textcolor{gray!90}{\scriptsize$\pm$#2}}

\resizebox{\textwidth}{!}{%
\begin{tabular}{@{}
c!{\hspace{\methodgap}}
l!{\hspace{\resultgap}}
c!{\hspace{\datasetgap}}c!{\hspace{\datasetgap}}c
!{\hspace{\labelgap}}
c!{\hspace{\datasetgap}}c!{\hspace{\datasetgap}}c
!{\hspace{\pretraingap}}
c!{\hspace{\datasetgap}}c!{\hspace{\datasetgap}}c
!{\hspace{\labelgap}}
c!{\hspace{\datasetgap}}c!{\hspace{\datasetgap}}c@{}}

\toprule
\multirow{3}{*}{\vspace{-1.2em}\shortstack{Size \\{(\%)}}} &
\multirow{3}{*}{Method} &
\multicolumn{6}{c}{{CIFAR-10 Pretraining}} &
\multicolumn{6}{c}{{CIFAR-100 Pretraining}} \\
\cmidrule(lr){3-8}\cmidrule(l){9-14}

& &
\multicolumn{3}{c}{1\% Labels} &
\multicolumn{3}{c}{5\% Labels} &
\multicolumn{3}{c}{1\% Labels} &
\multicolumn{3}{c}{5\% Labels} \\
\cmidrule(lr){3-5}\cmidrule(lr){6-8}
\cmidrule(lr){9-11}\cmidrule(l){12-14}

& &
C10 & C100 & TIN &
C10 & C100 & TIN &
C10 & C100 & TIN &
C10 & C100 & TIN \\
\midrule

100 & \textsc{Full Data}
& \acc{87.44}{0.56} & \acc{26.99}{1.20} & \acc{12.89}{0.26}
& \acc{89.76}{0.17} & \acc{41.77}{0.65} & \acc{22.45}{0.20}
& \acc{69.79}{0.57} & \acc{49.68}{0.82} & \acc{15.35}{0.33}
& \acc{76.79}{0.34} & \acc{58.96}{0.36} & \acc{25.70}{0.58} \\

\cmidrule(l){1-14}

\multirow{3}{*}{0.2} & \textsc{Random}
& \acc{39.92}{1.64} & \acc{10.40}{0.63} & \acc{4.93}{0.43}
& \acc{46.84}{2.21} & \acc{18.54}{0.83} & \acc{9.19}{0.73}
& \acc{37.67}{3.21} & \acc{10.03}{1.73} & \acc{4.51}{0.82}
& \acc{44.06}{3.90} & \acc{17.26}{2.61} & \acc{8.41}{1.39} \\

& \textsc{MKDT}
& \acc{36.38}{1.49} & \acc{8.56}{1.17} & \acc{4.00}{0.44}
& \acc{42.64}{3.19} & \acc{16.11}{1.19} & \acc{7.72}{0.49}
& \acc{37.52}{0.97} & \acc{10.18}{0.34} & \acc{4.58}{0.18}
& \acc{44.94}{0.48} & \acc{18.18}{0.33} & \acc{8.73}{0.16} \\

& \textbf{MIPA}
& \best{40.07}{0.78} & \best{10.43}{0.59} & \best{4.94}{0.18}
& \best{47.66}{0.84} & \best{18.70}{0.19} & \best{9.28}{0.25}
& \best{40.10}{0.86} & \best{12.00}{0.59} & \best{5.18}{0.22}
& \best{48.34}{0.98} & \best{20.30}{0.64} & \best{9.70}{0.36} \\

\arrayrulecolor{black!65}
\cmidrule[0.1pt](l){2-14}
\arrayrulecolor{black}

\multirow{3}{*}{1} & \textsc{Random}
& \acc{49.34}{1.10} & \acc{14.20}{0.47} & \best{7.20}{0.43}
& \acc{56.46}{0.90} & \acc{23.90}{0.88} & \acc{13.19}{0.41}
& \acc{46.47}{1.39} & \acc{15.09}{1.05} & \acc{7.17}{0.73}
& \acc{54.36}{1.36} & \acc{24.78}{1.10} & \acc{13.09}{0.99} \\

& \textsc{MKDT}
& \acc{43.32}{1.32} & \acc{11.60}{0.38} & \acc{5.86}{0.43}
& \acc{51.77}{0.77} & \acc{20.75}{0.87} & \acc{11.01}{0.42}
& \acc{42.41}{1.80} & \acc{12.03}{0.67} & \acc{6.01}{0.22}
& \acc{50.75}{1.34} & \acc{21.21}{0.48} & \acc{10.91}{0.36} \\

& \textbf{MIPA}
& \best{50.65}{1.25} & \best{14.23}{0.93} & \acc{7.18}{0.27}
& \best{57.67}{0.68} & \best{24.01}{0.24} & \best{13.46}{0.23}
& \best{47.67}{1.27} & \best{17.05}{0.86} & \best{7.63}{0.35}
& \best{56.38}{0.28} & \best{27.29}{0.40} & \best{14.04}{0.43} \\

\bottomrule
\end{tabular}
}
\end{table*}

\vspace{-1em}
\paragraph{Effect of initialization.}
We investigate how the initialization of synthetic images affects \textsc{MIPA}. We compare randomly selected real images, K-means, Invariant Information Clustering (IIC)~\cite{9008122}, and Gaussian noise. Since optimization from noise may require more iterations to converge, we also evaluate Gaussian initialization with $3\times$ and $10\times$ the default distillation iterations. Table~\ref{tab:mipa-initialization-matched} presents the results.

Random real-image initialization achieves the highest mean accuracy across all settings, with neither clustering method providing an advantage. One possible explanation is that clustering prioritizes representative samples based on feature-space structure, which may not necessarily align with the augmentation-invariant information targeted by \textsc{MIPA}. In contrast, random sampling provides diverse real-image structures without introducing clustering bias. Gaussian noise initialization performs substantially worse, even with $10\times$ more iterations. Although additional iterations partially improve performance, the gains are small and inconsistent, and accuracy remains below all real-image-based initializations. These findings suggest that real images provide useful initial structure that is difficult to recover from noise, while random sampling offers an effective initialization without additional clustering overhead.

\begin{table}[h]
\centering
\caption{Effect of \textsc{MIPA} initialization on CIFAR-10 linear probing accuracy (\%, mean $\pm$ std.\ over five runs). Iteration multipliers apply to Gaussian noise initialization. Bold marks the best result.}
\vspace{-0.6em}
\label{tab:mipa-initialization-matched}
\mipatable
\setlength{\tabcolsep}{1.1pt}
\renewcommand{\arraystretch}{1.021}
\begin{tabular*}{\columnwidth}{@{\extracolsep{\fill}}l*{4}{r}@{}}
\toprule
\multirow{2}{*}[-2.5pt]{Initialization} & \multicolumn{2}{c}{\textbf{0.2\% data}} & \multicolumn{2}{c}{\textbf{1\% data}} \\
\cmidrule(lr){2-3}\cmidrule(l){4-5}
 & 1\% labels & 5\% labels & 1\% labels & 5\% labels \\
\midrule
{Random} & \mipabest{45.18}{0.41} & \mipabest{50.58}{0.47} & \mipabest{52.26}{0.55} & \mipabest{57.36}{0.16} \\
K-means & \mipaacc{43.72}{0.28} & \mipaacc{49.52}{0.38} & \mipaacc{50.73}{0.98} & \mipaacc{56.60}{0.23} \\
IIC & \mipaacc{43.32}{0.54} & \mipaacc{49.59}{0.35} & \mipaacc{50.15}{0.51} & \mipaacc{55.91}{0.14} \\
\midrule
Gaussian noise & \mipaacc{39.18}{0.74} & \mipaacc{44.62}{0.72} & \mipaacc{38.13}{0.71} & \mipaacc{44.00}{0.37} \\
\quad $3\times$ iter. & \mipaacc{41.98}{0.56} & \mipaacc{47.74}{0.37} & \mipaacc{42.82}{0.73} & \mipaacc{48.02}{0.30} \\
\quad $10\times$ iter. & \mipaacc{42.62}{0.36} & \mipaacc{47.48}{0.32} & \mipaacc{43.13}{0.30} & \mipaacc{48.82}{0.34} \\
\bottomrule
\end{tabular*}
\end{table}

\vspace{-0.8em}
\paragraph{Computational cost.}
A key advantage of our method is its computational efficiency for distillation.
Table~\ref{tab:generation-cost} compares the distillation runtime and disk storage requirements of \textsc{MKDT} and \textsc{MIPA}, excluding SSL teacher pretraining, which is required by our method and both \textsc{MKDT} and \textsc{KRR-ST}. All runtimes were measured on a single NVIDIA L40S GPU.
These measurements quantify distillation overhead rather than total end-to-end training cost. On CIFAR-10 and CIFAR-100, \textsc{MIPA} is approximately $5.8$--$8.0\times$ faster than \textsc{MKDT}, even before accounting for the additional cost of generating \textsc{MKDT}'s expert trajectories. On Tiny ImageNet, \textsc{MIPA} remains faster, although the speedup decreases as the synthetic dataset size increases. Furthermore, by eliminating the need for expert trajectory buffers, \textsc{MIPA} reduces intermediate disk storage by more than $250\times$ across all three datasets, substantially lowering the resources required for distillation.

\begin{table}[h]
\centering
\caption{Distillation costs excluding teacher training. Image optimization time is reported by synthetic budget, with expert trajectory generation measured separately. Storage covers intermediate distillation data.}
\vspace{-0.6em}
\label{tab:generation-cost}
\mipatable
\setlength{\tabcolsep}{1.4pt}
\renewcommand{\arraystretch}{1.0}
\begin{tabular*}{\columnwidth}{@{\extracolsep{\fill}}ll*{5}{r}@{}}
\toprule
\multirow{2}{*}[-2.5pt]{Dataset} & \multirow{2}{*}[-2.5pt]{Method} & \multicolumn{3}{c}{Time (min)} & \multirow{2}{*}[-2.5pt]{\shortstack{Experts\\(GPU-h)}} & \multirow{2}{*}[-2.5pt]{\shortstack{Storage\\(GiB)}} \\
\cmidrule(lr){3-5}
 &  & 0.2\% & 1\% & 2\% &  &  \\
\midrule

\multirow{2}{*}{CIFAR-10} & \textsc{MKDT} & 47.82 & 112.90 & 112.83 & 1.01 & 10.65 \\
& \textsc{MIPA} & \textbf{6.43} & \textbf{14.07} & \textbf{19.48} & -- & \textbf{0.042} \\
\midrule
\multirow{2}{*}{CIFAR-100} & \textsc{MKDT} & 47.98 & 112.95 & 114.40 & 0.98 & 10.65 \\
& \textsc{MIPA} & \textbf{6.38} & \textbf{14.12} & \textbf{19.50} & -- & \textbf{0.042} \\
\midrule
\multirow{2}{*}{\shortstack[l]{Tiny\\ImageNet}} & \textsc{MKDT} & 200.62 & 232.43 & 230.62 & 7.13 & 11.91 \\
& \textsc{MIPA} & \textbf{39.15} & \textbf{128.58} & \textbf{192.68} & -- & \textbf{0.042} \\
\bottomrule
\end{tabular*}
\end{table}

\begin{figure}[h]
\centering
\begingroup
\setlength{\tabcolsep}{0.5pt}
\renewcommand{\arraystretch}{0}
\begin{tabular}{@{}*{10}{c}@{}}
\includegraphics[width=0.095\columnwidth]{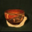} &
\includegraphics[width=0.095\columnwidth]{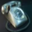} &
\includegraphics[width=0.095\columnwidth]{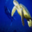} &
\includegraphics[width=0.095\columnwidth]{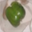} &
\includegraphics[width=0.095\columnwidth]{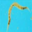} &
\includegraphics[width=0.095\columnwidth]{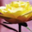} &
\includegraphics[width=0.095\columnwidth]{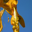} &
\includegraphics[width=0.095\columnwidth]{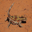} &
\includegraphics[width=0.095\columnwidth]{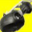} &
\includegraphics[width=0.095\columnwidth]{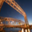} \\[2pt]
\includegraphics[width=0.095\columnwidth]{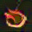} &
\includegraphics[width=0.095\columnwidth]{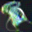} &
\includegraphics[width=0.095\columnwidth]{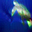} &
\includegraphics[width=0.095\columnwidth]{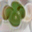} &
\includegraphics[width=0.095\columnwidth]{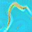} &
\includegraphics[width=0.095\columnwidth]{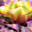} &
\includegraphics[width=0.095\columnwidth]{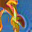} &
\includegraphics[width=0.095\columnwidth]{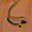} &
\includegraphics[width=0.095\columnwidth]{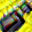} &
\includegraphics[width=0.095\columnwidth]{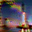} \\[2pt]
\includegraphics[width=0.095\columnwidth]{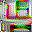} &
\includegraphics[width=0.095\columnwidth]{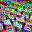} &
\includegraphics[width=0.095\columnwidth]{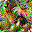} &
\includegraphics[width=0.095\columnwidth]{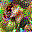} &
\includegraphics[width=0.095\columnwidth]{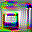} &
\includegraphics[width=0.095\columnwidth]{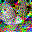} &
\includegraphics[width=0.095\columnwidth]{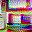} &
\includegraphics[width=0.095\columnwidth]{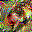} &
\includegraphics[width=0.095\columnwidth]{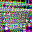} &
\includegraphics[width=0.095\columnwidth]{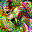}
\end{tabular}
\endgroup
\caption{CIFAR-100 prototype examples. Top: original training images used for initialization at the 1\% data budget. Middle: corresponding prototypes after 8k \textsc{MIPA} iterations. Bottom: independent Gaussian-noise-initialized prototypes at the 0.2\% data budget after 50k iterations.
}
\label{fig:cifar100-prototype-evolution}
\end{figure}

\paragraph{Prototype visualization.}
Figure~\ref{fig:cifar100-prototype-evolution} illustrates the synthetic images generated by \textsc{MIPA} using real-image initialization, alongside examples initialized from Gaussian noise. With real-image initialization, the optimized CIFAR-100 prototypes largely preserve the original object structures while showing noticeable changes in color, contrast, and texture. Some examples exhibit more distinct boundaries, while others develop visible distortions, indicating that \textsc{MIPA} modifies the original images during distillation. 
In contrast, the Gaussian-noise-initialized prototypes exhibit geometric patterns and local textures but little recognizable object structure.

\section{Conclusion}



In this work, we introduced \textsc{MIPA}, a self-supervised dataset distillation method that enables SSL pretraining from scratch using only synthetic images, 
without requiring auxiliary target representations during SSL pretraining.
By maximizing mutual information between cross-view prototype assignments, \textsc{MIPA} produces compact synthetic datasets that outperform the evaluated image-only baselines while substantially reducing distillation time and storage requirements. Extensive evaluations on image classification and various downstream tasks demonstrate the effectiveness of \textsc{MIPA} across different model architectures, SSL methods, and synthetic dataset sizes.



{
    \small
    \bibliographystyle{ieeenat_fullname}
    \bibliography{ref}
    
}


\end{document}